\documentclass[letterpaper, 10 pt, conference]{ieeeconf}  

\IEEEoverridecommandlockouts                               
                                                          
\usepackage{cite}
\usepackage{amsmath,amssymb,amsfonts}
\usepackage{algorithmic}
\usepackage{graphicx}
\usepackage{textcomp}
\usepackage{xcolor}
\usepackage{tablefootnote}
\usepackage[caption=false, font=footnotesize]{subfig}
\usepackage{flushend}
\usepackage{gensymb}
\usepackage{tabularx}
\usepackage{lipsum}
\usepackage{multirow}
\usepackage{hyperref}
\usepackage[british]{babel}
\usepackage{tikz}

\title{\LARGE \bf
A Tilt-Rotor UAV with a Gripper for Stable Contact-Based Tasks via Environmental Anchoring
}

\author{Joshua Taylor$^{1,2}$, Nursultan Imanberdiyev$^{2}$, Wei-Yun Yau$^{2}$, Guillaume Sartoretti$^{1}$, and Efe Camci$^{2}$ %
\thanks{$^{1}$National University of Singapore (NUS), Singapore.}%
\thanks{$^{2}$Institute for Infocomm Research (I$^2$R), A*STAR, Singapore.
}%
}

\begin{document}

\maketitle
\thispagestyle{empty}
\pagestyle{empty}

\begin{abstract}

Maintaining a stable pose during physical interaction is a significant challenge for aerial robots, often limiting their use in contact-based tasks. This paper presents a novel uncrewed aerial vehicle (UAV) platform designed to transition from unconstrained flight to a stable, constrained work platform via environmental anchoring. Our system comprises: 1) a multirotor with a tilt-rotor mechanism that decouples pitch from forward motion, enabling stable hover at non-zero pitch angles, and 2) a novel underactuated, cable-driven, prismatic gripper featuring compliance to adapt to irregular geometries, designed to stabilize the UAV by anchoring it to its environment. 
We present the design and prototyping of the complete system and validate its performance through a series of real-robot flight tests. 
Results demonstrate that anchoring significantly improves stability for interaction tasks, reducing positional drift RMSE by over 95\% compared to a free-flight baseline, even under windy conditions. The anchored system can withstand longitudinal reaction forces up to 75\,N while maintaining a stable pose. Furthermore, across a range of target geometries and orientations, the system demonstrated consistent stability with a positional drift RMSE that never exceeded 3\,mm. These results establish the viability of our approach for complex physical interaction tasks, such as sampling tree health by drilling or sensor installation in hard-to-reach locations. Watch our UAV at: \url{https://youtu.be/HDQ8S4ZW3Ls}.

\end{abstract}

\section{Introduction}

Uncrewed aerial vehicles (UAVs) are increasingly used for contact-based tasks in hard-to-reach locations~\cite{dautzenberg2023perching,bodie2020active}. A particularly challenging research direction is deploying aerial robots in natural environments, such as for forest monitoring~\cite{aucone2023drone}. There, tasks like drilling into branches to assess tree health require the robot to exert controlled forces with high pose stability.

This presents two key challenges for aerial manipulation. First, \textit{versatility}: the environment is unstructured, with irregular, rough branches at arbitrary angles.
Second, \textit{stability}: physical interaction often requires high stability to minimize damage to the tool and the environment, and introduces external forces that can destabilize a hovering platform.

Current aerial platforms often fail to meet both requirements simultaneously. While advanced tilt-rotor designs can generate forces at various orientations~\cite{hameed2025dragonfly,lee2023minimally}, they typically lack suitable end-effectors for robustly anchoring to natural surfaces. Conversely, while perching mechanisms exist, they are often designed for specific conditions such as horizontal branches~\cite{aucone2023drone} or smooth vertical walls~\cite{dautzenberg2023perching}, or they require aggressive maneuvers to engage~\cite{askari2024crash}. These mechanisms may not provide the necessary stability or versatility for performing subsequent high-force tasks on the irregular, inclined surfaces found in natural environments.

\begin{figure}[t!]
    \centering
    \includegraphics[width=0.7\columnwidth]{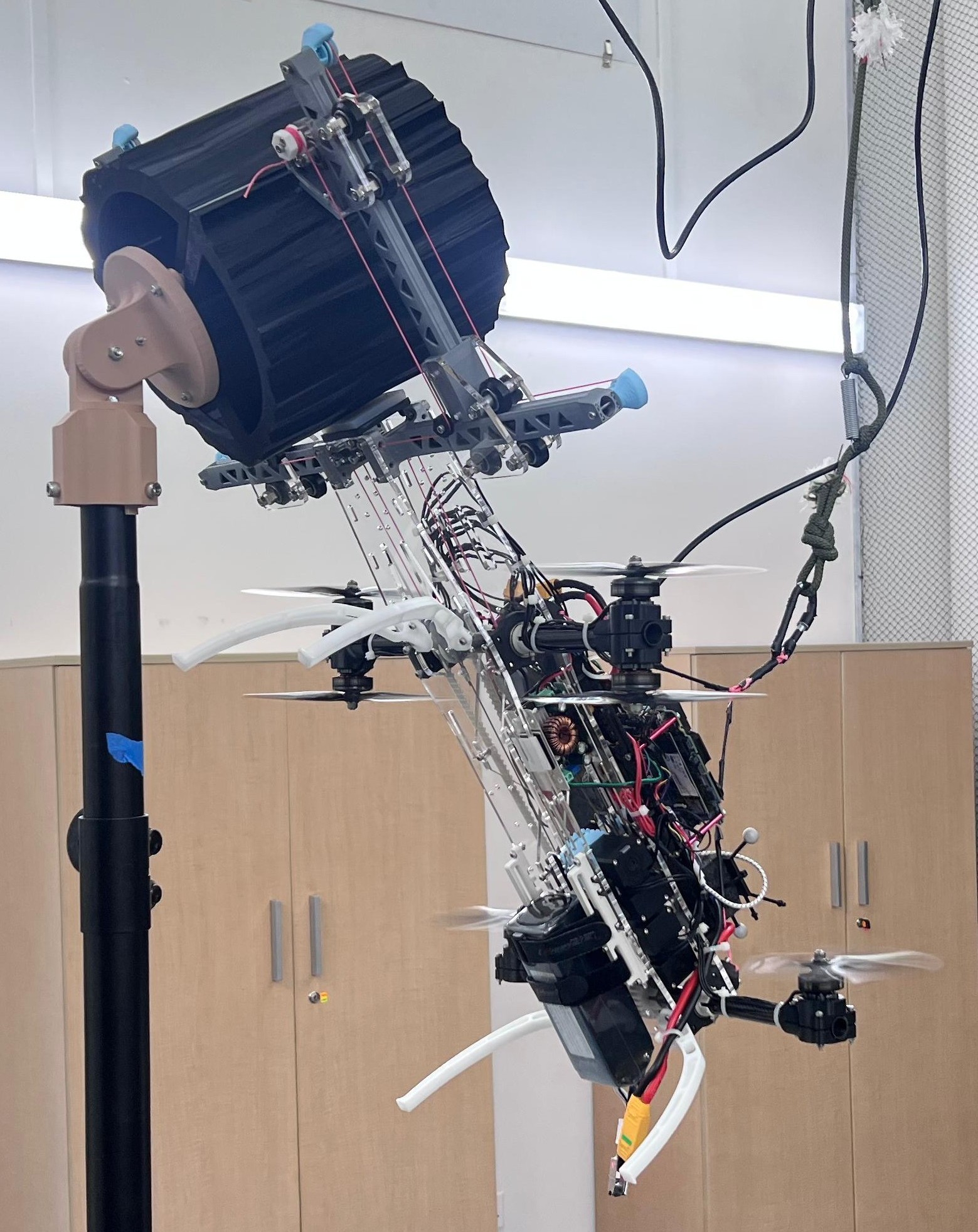}
    \caption{Tilt-rotor UAV with cable-driven gripper anchored onto a 60\textdegree~overhanging \textit{tree branch} to establish a stable work platform.}
    \vspace{-0.5cm}
    \label{fig_front_page}
\end{figure}

We present an aerial robot to address the versatility and stability required for contact-based tasks in natural environments. Our robot integrates a tilt-rotor UAV with a novel cable-driven prismatic gripper. The tilt-rotors enable the UAV to hover perpendicularly to inclined branches, and the gripper allows for robust anchoring to irregular targets. We demonstrate that via anchoring, our robot achieves stability suitable for complex physical interaction tasks such as drilling, even under difficult flight conditions like windy environments and large-angle pitching, as shown in Fig.~\ref{fig_front_page}. 

The main contributions of our work are summarized as:
\begin{enumerate}
    \item The design and build of a novel cable-driven, prismatic gripper for grasping a diverse range of targets.
    \item Integration of the gripper into a tilt-rotor platform to achieve environmental anchoring.
    \item Validation of the performance improvements gained by anchoring under various flight conditions.
    \item Quantification of the gripper's limits under high loads and its versatility across various target geometries.
\end{enumerate}

The remainder of this paper is structured as follows. Section~\ref{sec:related_work} discusses the literature on aerial physical interaction, Sec.~\ref{sec_the_uav} introduces the UAV and gripper design, Sec.~\ref{sec_control} introduces the control scheme used, and Sec.~\ref{sec_testing} describes the experiments performed and the results obtained. Finally, Sec.~\ref{sec_conclusion} provides key conclusions from our work.

\section{Related Work}
\label{sec:related_work}

To perform contact-based tasks, an aerial robot must be able to generate forces at arbitrary orientations to align with the target and ensure high pose stability during interaction. We review the state of the art in these two areas.

\subsection{Force Generation for Aerial Interaction}

Conventional underactuated multirotors are limited by coupled translational and rotational dynamics, which restricts their ability to apply sustained forces against non-horizontal surfaces~\cite{scholten2013interaction}. Conversely, fully- and over-actuated platforms can generate forces and torques in all 6-degrees of freedom (DoF), often by employing individually tiltable rotors~\cite{hameed2025dragonfly}. These designs enable capabilities like active force control for contact-based inspection~\cite{bodie2020active}. However, the required mechanical complexity (e.g., additional actuators, complex linkages) increases weight and power consumption, thereby limiting flight endurance and payload capacity. Consequently, researchers have explored UAV designs with 5 controlled DoFs (CDoF) as a practical alternative~\cite{lee2023minimally, ding2021design}. Typically requiring only one or two more actuators, such designs decouple longitudinal motion from the pitch axes, allowing the maneuverability needed to approach and interact with inclined surfaces without the excessive weight of 6-CDoF architectures. This balances performance and complexity for contact-based tasks like aerial drilling~\cite{ding2021design} or perching.

\subsection{Perching and Grasping on Unstructured Surfaces}

Leveraging the environment for support through perching can enhance stability and force exertion during interaction. The work in~\cite{hang2019perching} showed that such support can overcome limitations of free-flight interaction, including high energy consumption and limited pose stability. However, the effectiveness of this approach is dependent on the perching mechanism's ability to interface with the target surface. On structured and smooth surfaces, suction cups have been used effectively for high-force tasks such as drilling~\cite{dautzenberg2023perching}. However, such methods typically fail on rough, porous, or irregular surfaces common in natural environments.

For perching onto branches, grippers with revolute joints are widely used. Designs such as~\cite{roderick2021bird} mimic avian claws to encircle and secure branches. The review in~\cite{meng2022aerial} indicates that most UAV-mounted grippers utilize similar multi-fingered revolute designs. While underactuation can enable compliance, this architecture is better suited to smaller, cylindrical shapes. Furthermore, these designs are typically mounted beneath the airframe, relying on agile or aggressive maneuvers to engage the target~\cite{askari2024crash,zheng2024albero}. Such approaches present a risk for heavier platforms or operations in cluttered spaces. The work in~\cite{taylor2024reconfigurable} used a front-mounted revolute gripper to improve pose stability against vertical branches, but found it less adept at forming a stable grasp on larger or unstructured surfaces. Generally, revolute grippers also suffer from grasp dynamics that are inherently sensitive to the target’s geometry and the joint angles during contact~\cite{balasubramanian2011comparison}. Furthermore, the fingers can generate unintended outward forces during the initial arcing movement of the grasp~\cite{backus2018prismatic}. While we can look to concepts like reconfiguration through extendable fingers~\cite{wang2024robot} to address grasping of a broader range of target shapes and sizes, they ultimately still suffer from these same undesired grasp behaviors. Alternatively, continuum-style grippers~\cite{peng2025dexterous} offer strong compliance but typically lack the rigidity for a robust anchor.

Conversely, prismatic grippers with parallel jaws provide predictable clamping forces independent of the target object size. The benefits of this architecture were demonstrated in the hybrid prismatic-revolute gripper in~\cite{backus2018prismatic}, suggesting that prismatic joints offer a strong direction for establishing stable work platforms in unstructured environments. The broader literature on prismatic grippers explores relevant areas like adaptability~\cite{lee2025frictional}, as well as maximizing stroke length~\cite{kobayashi2019design} and force output~\cite{takaki2007grasp}, but lacks a lightweight design capable of complying with a large range of target shapes and sizes while withstanding large interaction forces.

Overall, the literature highlights a need for a system capable of perching on irregular, pole-like objects at various angles without aggressive maneuvers. Such a need could be fulfilled by the co-design of a capable aerial platform and a gripper with both predictable dynamics and compliance.

\begin{figure*}
  \includegraphics[width=0.9\textwidth]{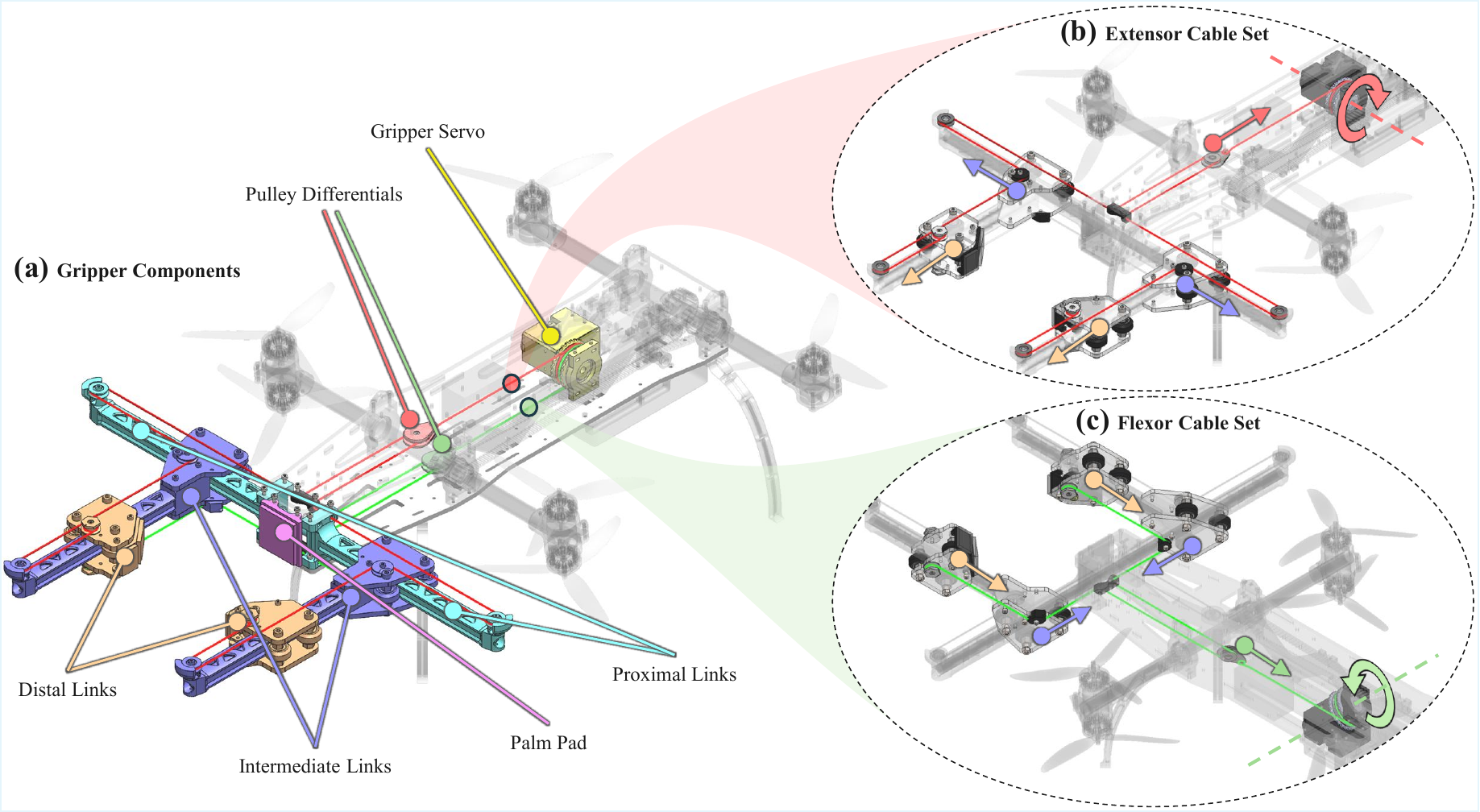}
  \centering
  \vspace{-0.3cm}
  \caption{The gripper design and cable actuation. \textbf{(a)} Key mechanical components. \textbf{(b)} Routing of the extensor cable set showing how rotation of the gripper servo translates into opening the gripper. \textbf{(c)} Routing of the flexor cable set, viewed from the underside of the UAV, showing how opposite rotation of the gripper servo translates into closure of the gripper.}
  \vspace{-0.35cm}
  \label{fig_gripper_labeled}
\end{figure*}

\section{UAV Design}
\label{sec_the_uav}

To address the need for a stable and versatile UAV for contact-inspection tasks, we designed and built a tilt-rotor UAV with a gripper. While the tilt-rotors enable decoupling of forward motion from pitching, allowing the UAV to hover at pitch angles of up to -60\textdegree{} (nose up), the gripper enables robust anchoring against a range of target objects, including cylindrical, rectangular, and irregularly shaped columns and poles. This combination of capabilities is particularly beneficial for inspection work in natural environments, where, for example, tree branches of varying cross-sectional diameters, geometries, and angles can be found. The gripper is deliberately mounted to the front of the UAV to enable mounting of long-bodied instruments, such as the Resistograph\textregistered~\cite{rinntech_resistograph} used for internal wood inspection. The remainder of this section describes the design and build of our UAV.

\subsection{Airframe}

Our UAV is based on a 5-CDoF H-frame chassis, similar to that in~\cite{lee2023minimally}. The front-mounted gripper generates a constant pitch torque, which we compensate for with an asymmetric rotor configuration. Two coaxial sets of rotors are positioned on the front cross-arm and two single rotors on the parallel rear arm. All rotors tilt simultaneously using a timing belt system driven by a servo motor. This servo is placed towards the back of the UAV to further offset the gripper mass. The axis of rotation for the tilting mechanism is aligned with the longitudinal axis of each rotor arm. This configuration provides an additional degree of control, as tilting the rotors generates a net thrust vector in the UAV's body-$x$ direction, allowing for decoupled pitch and forward motion.

\begin{figure}[h!]
    \centering
    \includegraphics[width=0.9\columnwidth]{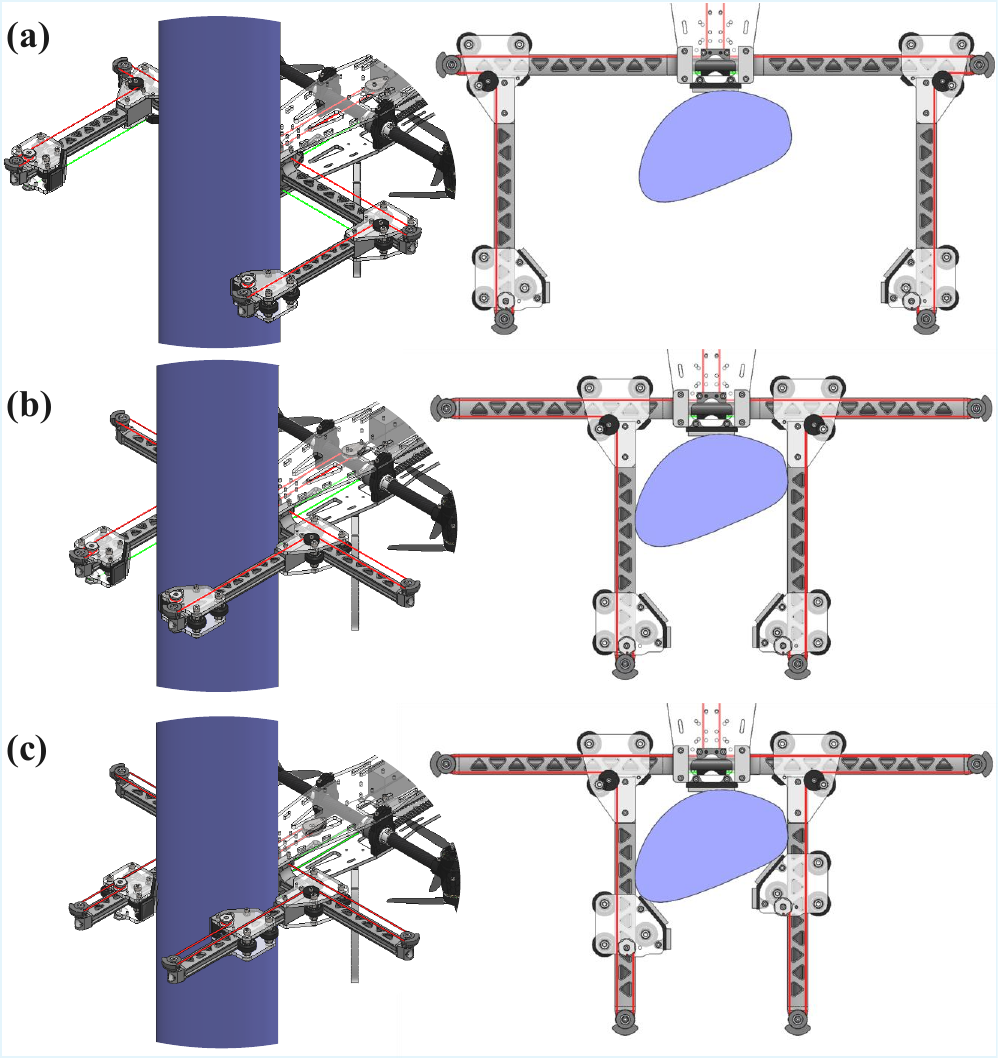}
    \caption{An example sequence of the gripper grasping onto an irregularly shaped target object, including an isometric and top-down view of the scene. The gripper starts in the \textbf{(a)} open position. As the servo applies tension to the flexor cable set, the intermediate links move preferentially to the distal links due to their differing joint stiffnesses, resulting in the position shown in \textbf{(b)}. Once the intermediate links cannot close any further, the distal links form the \textbf{(c)} completed grasp, complying with the target object.}
    \vspace{-0.35cm}
    \label{fig_gripper_closing}
\end{figure}

\subsection{Gripper}

We present our gripper for physical interaction. The design has two fingers, each consisting of two prismatic joints. As shown in Fig.~\ref{fig_gripper_labeled}(a), each finger has a proximal link, which acts as a linear rail for an intermediate link, which in turn acts as a linear rail for the distal link. This configuration provides both a clamping force on the target object (between the intermediate links) and a pulling force that draws the target and the UAV together (due to the distal links). The gripper is underactuated, in that a single servo motor drives all four prismatic joints. The actuation is transmitted via a pair of antagonistic cable sets, highlighted in Fig.~\ref{fig_gripper_labeled}(b) and~\ref{fig_gripper_labeled}(c). Both sets are connected to a servo-mounted cable drum and terminate at the distal links. Rotating the servo in one direction tensions the extensor cable set, which opens the jaws. Opposite tensions the flexor set, applying a closing and clamping force. This cable-driven approach enables us to position a portion of the gripper mass (i.e., the servo) at the back end of the UAV, thereby reducing weight at the extremities and contributing to improved mass distribution. Each cable set runs through a set of pulleys and a pulley differential, allowing the separate joints to move independently of one another. This enables a degree of compliance when grasping irregularly shaped or off-center objects. Note that by adjusting the tension on the carriages that run along the linear rails, we can fine-tune the sequence in which the gripper closes. For example, a tighter carriage on the distal link with respect to the carriage on the intermediate link will cause the intermediate link to move preferentially over the distal link when closing the gripper. This not only maximizes the possible workspace of the gripper by ensuring it can utilize the full extension of the distal joints, but also allows the gripper to function at different pitch angles when the end effectors are subject to gravity in the direction of the joint. Figure~\ref{fig_gripper_closing} demonstrates the preferential movement of the intermediate links and the gripper's compliant nature.

Our design offers three advantages over the more common revolute-joint grippers. First, the linear caging motion pulls the UAV towards the static target, eliminating the \textit{push-away} effect of arcing revolute fingers and ensuring consistent UAV placement relative to the target. Second, the rectilinear workspace allows for robust interfacing with diverse geometries, including flat or rectangular objects, that are poorly matched to the crescent-shaped workspaces of revolute designs. Finally, clamping forces remain co-linear with the joint axes throughout the grasp, providing a consistent force vector and mechanical advantage. This ensures predictable dynamics and simplifies force control, in contrast to the angle-dependent forces inherent in rotational architectures.

\subsection{Prototype}

\begin{figure}[t!]
    \centering
    \includegraphics[width=0.8\columnwidth]{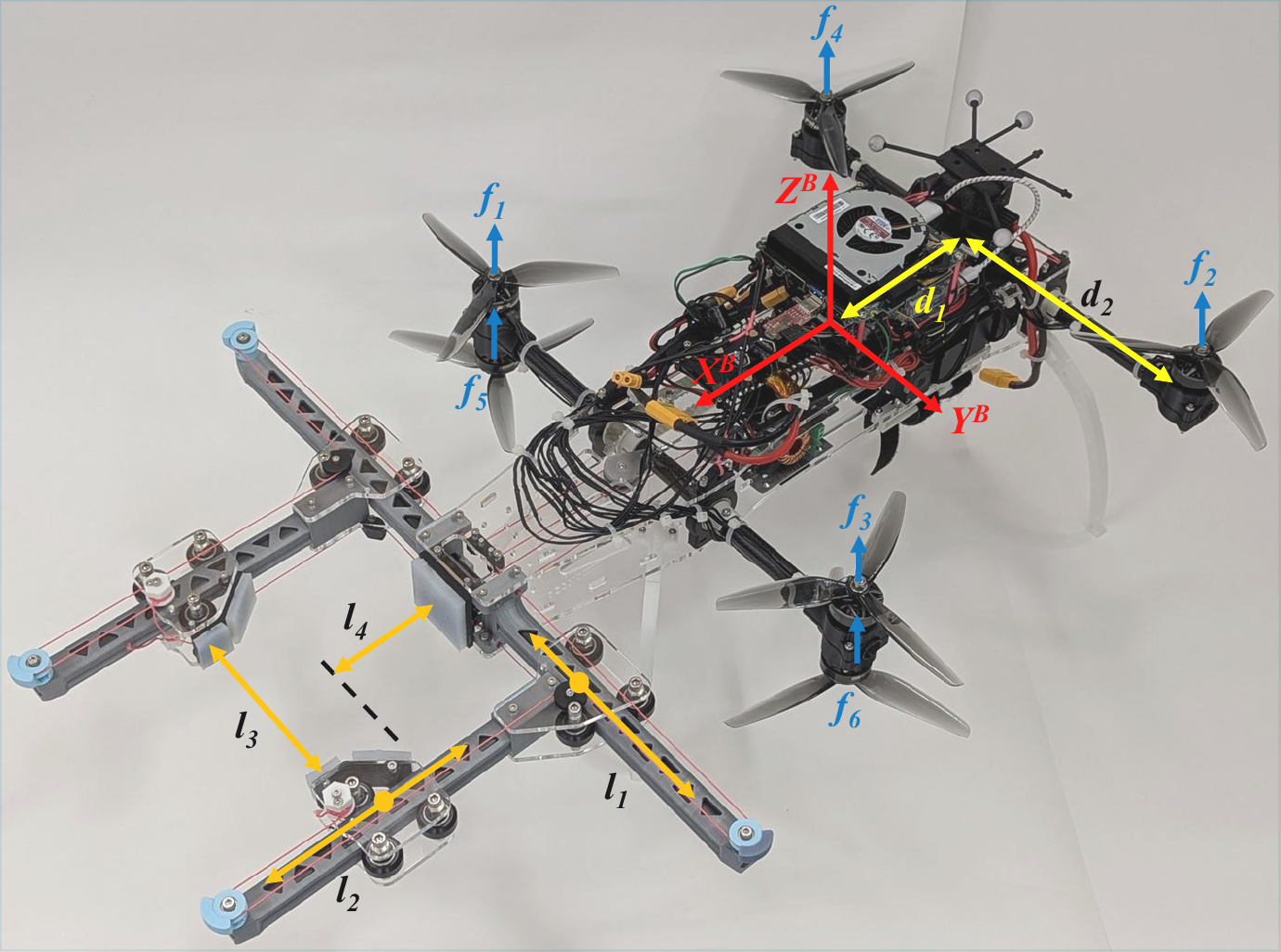}
    \caption{The prototype of our UAV with gripper, with the body frame axes ($X^B, Y^B, Z^B$), individual rotor thrusts ($f_*$), and key dimensions of the airframe ($d_*$) and gripper ($l_*$) labeled.}
    \vspace{-0.5cm}
    \label{fig_prototype_labeled}
\end{figure}

For real-robot experiments, we built a working prototype of the tilt-rotor UAV with the gripper, shown in Fig.~\ref{fig_prototype_labeled}. Table~\ref{tab:components} shows the key electrical components used, and Table~\ref{tab:dimensions} shows the key characteristics and dimensions of the system.

The airframe is fabricated from laser-cut acrylic, with carbon fiber tubes for the rotor arms. Custom structural components, gears, and pulleys were 3D-printed, primarily in Polylactic Acid (PLA), with Acrylonitrile Butadiene Styrene (ABS) used for heat-sensitive components such as rotor mounts. The tilt-rotor mechanism uses a T5 timing belt.

The gripper's assembly also leverages a combination of materials. Its linear rails are composed of 3D-printed PLA, reinforced internally with carbon fiber tubes for lightweight rigidity. The carriages, constructed from a mix of 3D-printed PLA and laser-cut acrylic, slide along these rails on roller bearings. Each carriage is equipped with three alternating roller bearings mounted on eccentric nuts, which allow for tuning of each joint's stiffness. The cable-driven actuation relies on a hybrid cable system: a 1-mm steel cable, chosen for its high tensile stiffness, is used for sections closer to the UAV's core, while a lighter 1-mm Dyneema line is employed for the gripper itself. Where the cables undergo a full 180\textdegree~turn, we run the cables around bearing-mounted pulleys to minimize cable friction. At the distal links, the cables terminate at adjustable 3D-printed fixtures, allowing for individual tensioning of each cable set. Finally, all contact surfaces on the palm and distal links are covered in a 6-mm layer of silicone rubber, providing increased friction, compliance with target objects, and shock absorption.

\section{Control}
\label{sec_control}

In this section, we introduce the control strategies used for the gripper and the UAV.

\subsection{Gripper Control}

We use the gripper servo's current reading to control the grasp. Specifically, once a grasp is initiated, the gripper continues to close until the current exceeds a pre-defined threshold. At this moment, denoted by $t_g$, the gripper stops and maintains its position for the duration of the grasp. The threshold was determined empirically to provide a secure grasp without damaging the target object and the gripper.

\begin{table}[t]
\caption{Key flight components used on the UAV prototype.}
\label{tab:components}
\centering
\begin{tabular}{@{}ll@{}}
\hline
\textbf{Component} & \textbf{Specification} \\
\hline
Flight controller & Pixhawk~6X \\
Electronic speed controller (ESC) & T-Motor FPV C-55A-8S-8IN1 \\
Motors & T-Motor F90 2806.5 1300~KV \\
Propellers & HQ~7\texttimes4\texttimes3 \\
Tilt servo & Dynamixel XH540-W150-T\\
Gripper servo & Dynamixel XM540-W270-T\\
Onboard computer & Intel NUC i7 \\
Battery & 6S 6500mAh \\
\hline
\end{tabular}
\end{table}

\begin{table}[t]
\caption{Key characteristics and dimensions of the system.}
\label{tab:dimensions}
\centering
\begin{tabular}{@{}ll@{}}
\hline
\textbf{Characteristic} & \textbf{Value} \\
\hline
\multicolumn{2}{l}{\textbf{\textit{Overall System}}} \\
\quad Total mass & 4.5\,kg \\
\quad Longitudinal moment arm, $d_1$ & 165\,mm \\
\quad Lateral moment arm, $d_2$ & 185\,mm \\
\multicolumn{2}{l}{\textbf{\textit{Gripper}}} \\
\quad Gripper mass (including servo) & 1\,kg \\
\quad Intermediate link range, $l_1$ & 145\,mm \\
\quad Distal link range, $l_2$ & 120\,mm \\
\quad Minimum distal-distal link separation, $l_{3,min}$ & 35\,mm \\
\quad Minimum distal-palm pad separation, $l_{4,min}$ & 25\,mm \\
\hline
\end{tabular}
\end{table}

\subsection{UAV Control}

\begin{figure}[t]
    \centering
    \includegraphics[width=0.9\columnwidth]{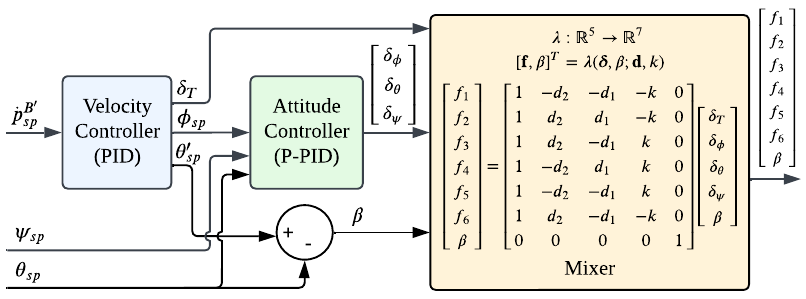}
    \caption{Our modified PX4 architecture with the UAV's pitch angle as the fifth control input.}
    \vspace{-0.5cm}
    \label{fig_control_diagram}
\end{figure}

Our UAV controller builds on the PX4~\cite{meier2015px4} cascaded PID control architecture. Following~\cite{chen2025simulation}, we modify the PX4 architecture to incorporate the tilt servo and include pitch as a fifth control input. The modified portion of the architecture is shown in Fig.~\ref{fig_control_diagram}, where $\dot{p}^{B'}_{sp}$ is the velocity setpoint in the $B'$ frame ($B$ frame leveled with respect to the horizon) generated by a position controller, and $\theta_{sp}$, $\phi_{sp}$, and  $\psi_{sp}$ are the pitch, roll, and yaw setpoints, respectively. The terms $\delta_T, \delta_\phi, \delta_\theta,$ and $\delta_\psi$ are the scalar thrust and torque inputs to the custom mixer matrix for our airframe. This matrix is defined by the airframe's geometric parameters, $\mathbf{d} = [d_1, d_2]^T$, and the motor drag-to-thrust coefficient, $k$. The control law for the tilt servo angle, $\beta$, is:
\begin{equation}
    \beta(t)=K_{p,\beta}e_v(t) + K_{i,\beta}\int_{0}^{t}e_v(\tau)d\tau + K_{d,\beta}\frac{de_v(t)}{dt}-\theta_{sp}(t)
\end{equation}
where $K_{p,\beta}$, $K_{i,\beta}$, and $K_{d,\beta}$ are the proportional, integral, and derivative (PID) gains of the velocity controller along $X^{B'}$ axis, and $e_v(t)$ is the velocity error in $X^{B'}$ at time $t$. Note that $\beta=0$ gives pure vertical thrust.

This architecture enables the UAV to hover at non-zero pitch angles. Our flight tests have validated a stable hover envelope from -60\textdegree{} (nose-up) to 55\textdegree{} (nose-down). The architecture also enables longitudinal motion through rotor tilting, providing a more stable approach for interaction.

However, as shown in~\cite{taylor2024reconfigurable}, position control with a fixed setpoint is ill-suited for physical interaction. The issue arises when the UAV transitions from free flight to a constrained state upon anchoring. If there is a mismatch between the setpoint and the current position, due to an existing position error or a displacement from the gripper pulling the UAV toward the target, conflict arises between the controller and the gripper. This leads to high power consumption, oscillations, instability, or even grasp failure. To mitigate this, we supplement the control architecture with a dynamic position setpoint. The position setpoint $p_{sp}(t)$ is defined as a piecewise function dependent on $t_g$. Before grasping, the setpoint is the fixed task position $p_{task}$. When the grasp is complete, it is set to the UAV's current position $p(t_g)$:

\begin{equation}
p_{sp}(t) = 
\begin{cases} 
p_{task} & \text{if } t < t_g \\
p(t_g) & \text{if } t \ge t_g 
\end{cases}
\end{equation}
This drives the position error to zero, minimizing steady-state conflict between the controller and the physical constraint. While transient conflict remains during grasp formation, we found that its duration (typically $<$\,\,4\,s) is short enough to not induce instability or significant excess power consumption.

\section{Testing}
\label{sec_testing}

We validate our system's performance through real-robot flight tests. Our evaluation begins with characterizing the platform's stabilization across diverse flight and disturbance conditions in Sec~\ref{sec:stab_perf}. Subsequently, we quantify the gripper's mechanical limits and versatility across various target geometries in Sec.~\ref{sec:grip_quant}. Finally, we qualitatively assess end-effector precision for contact-based tasks using a felt-tip pen as a contact proxy in Sec.~\ref{sec:prec_app}. We use two UAV configurations to quantify the stability gains provided by anchoring relative to the limits of active flight control:
\begin{enumerate}
    \item \textbf{Free Flight:} A 3.67-kg version of our UAV with the gripper removed (excluding the gripper servo, which serves a structural purpose). This yields a baseline for typical UAV performance in interaction tasks. 
    \item \textbf{Grasping:} Our proposed system, where the UAV transitions to a constrained state via physical anchoring.
\end{enumerate}
Both configurations use individual controller gains with similar trial-and-error tuning effort. For the \textbf{Free Flight} configuration, station-keeping performance was comparable to that demonstrated in other works~\cite{hang2019perching}. We note that the coaxial rotor configuration for the front was retained for the \textbf{Free Flight} tests. While not an optimal configuration without the gripper mass to offset, there remains sufficient control headroom for stable flight in this configuration. At hover, throttle values are $\approx$\,60\%\,/\,50\% (front/rear) with the gripper mounted and $\approx$\,40\%\,/\,60\% with the gripper removed. To account for the UAV's asymmetry, the pitch- and roll-axis gains were independently tuned for both configurations.

In all experiments, we use an OptiTrack motion capture system for the UAV's state estimation, and we assume the target object's pose is fixed and known a priori.

\begin{table*}[t]
\centering
\caption{Summary of flight performance.  All values are reported as the mean $\pm$ 1 sd over five trials.}
\vspace{-0.2cm}
\label{tab:comprehensive_metrics}
\begin{tabular}{@{}lcccc@{}}
\hline
 & \multicolumn{2}{c}{\textbf{No Wind}} & \multicolumn{2}{c}{\textbf{With Wind}} \\
\textbf{Metric} & \textbf{Free Flight} & \textbf{Grasping} & \textbf{Free Flight} & \textbf{Grasping} \\
\hline
\multicolumn{5}{@{}l}{\textit{\textbf{Level Flight}}} \\
\quad 3D Positional Drift RMSE (m)       & 0.015 $\pm$ 0.003 & 0.002 $\pm$ 0.001 & 0.037 $\pm$ 0.011 & 0.002 $\pm$ 0.000  \\
\quad Max 3D Positional Drift (m)  & 0.027 $\pm$ 0.006 & 0.004 $\pm$ 0.001 & 0.062 $\pm$ 0.017 & 0.004 $\pm$ 0.001  \\
\quad 3D Angular RMSE ($^{\circ}$)        & 2.24 $\pm$ 0.56   & 0.32 $\pm$ 0.04   & 4.02 $\pm$ 1.94   & 0.32 $\pm$ 0.06    \\
\quad Max 3D Angular Drift ($^{\circ}$)   & 4.09 $\pm$ 0.98   & 0.62 $\pm$ 0.06   & 7.15 $\pm$ 2.74   & 0.64 $\pm$ 0.15    \\

\multicolumn{5}{@{}l}{\textit{\textbf{Pitched Flight}}} \\
\quad 3D Positional Drift RMSE (m)       & 0.028 $\pm$ 0.013 & 0.002 $\pm$ 0.001 & 0.046 $\pm$ 0.013 & 0.002 $\pm$ 0.001  \\
\quad Max 3D Positional Drift (m)  & 0.047 $\pm$ 0.016 & 0.004 $\pm$ 0.001 & 0.076 $\pm$ 0.021 & 0.005 $\pm$ 0.001  \\
\quad 3D Angular RMSE ($^{\circ}$)        & 2.93 $\pm$ 1.31   & 0.39 $\pm$ 0.05   & 3.71 $\pm$ 0.92   & 0.33 $\pm$ 0.02    \\
\quad Max 3D Angular Drift ($^{\circ}$)   & 5.24 $\pm$ 1.63   & 0.74 $\pm$ 0.10   & 7.46 $\pm$ 1.82   & 0.67 $\pm$ 0.08    \\
\hline
\vspace{-0.7cm}
\end{tabular}
\end{table*}

\subsection{Stabilization Performance}
\label{sec:stab_perf}

\begin{figure}[t!]
    \centering
    \includegraphics[width=0.9\columnwidth]{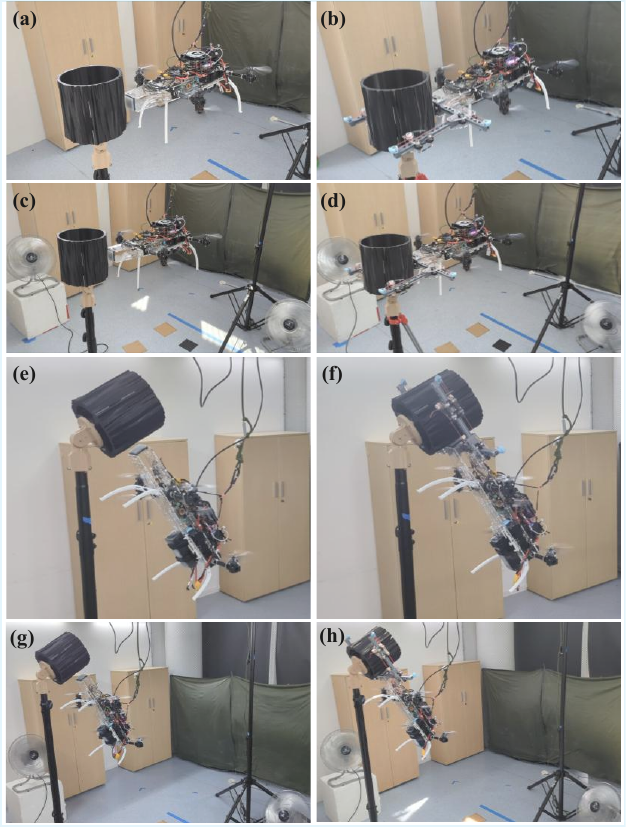}
    \vspace{-0.2cm}
    \caption{The eight experimental scenarios used to assess stabilization performance with \textbf{(b, d, f, h)} and without gripper \textbf{(a, c, e, g)}. They cover four primary conditions: \textbf{Level}: level flight without disturbance \textbf{(a, b)}; \textbf{Level + Wind}: level flight with wind \textbf{(c, d)}; \textbf{Pitched}: 60°-pitched flight without disturbance \textbf{(e, f)}; \textbf{Pitched + Wind}: 60°-pitched flight with wind \textbf{(g, h)}.}
    \vspace{-0.5cm}
    \label{fig_stabilization_tests}
\end{figure}

We evaluate the stabilization performance resulting from constraining the system using the gripper as an anchor. We refer to the following metrics for comparison:

\begin{enumerate}
    \item \textbf{3D Positional Drift RMSE:} The root mean square error of the time-series of 3D Euclidean distance (drift) values from the starting position.
    \item \textbf{Maximum 3D Positional Drift:} The maximum value in the time-series above.
    \item \textbf{3D Angular Drift RMSE:} The root mean square error of the time-series of the 3D angular distance (drift) from the starting orientation.
    \item \textbf{Maximum 3D Angular Drift:} The maximum value in the time-series above.
\end{enumerate}

We conducted tests under various flight conditions for both the \textbf{Free Flight} and \textbf{Grasping} configurations. These are:
\begin{enumerate}
    \item \textbf{Level}: Operation against a vertical target with no external disturbances.
    \item \textbf{Level + Wind}: Operation against a vertical target. There are wind disturbances created by two fans aimed at the level of the UAV. The fans are set to oscillation mode to create a varying disturbance.
    \item \textbf{Pitched}: Operation against a 60\textdegree-overhanging target with no external disturbances.
    \item \textbf{Pitched + Wind}: Operation against a 60\textdegree-overhanging target with the varying wind disturbance applied.
\end{enumerate}

Figure~\ref{fig_stabilization_tests} shows our testing scenarios. The chosen target object is a 3D-printed target object with a diameter of approximately 200\,mm and a textured surface to mimic tree bark. For these tests, $p_{task}$ is chosen such that the front of the UAV is approximately 2\,cm away from the target surface.

While the actual operation duration may vary based on the task of interest (e.g., for resistograph-based inspection, based on the size of the tree and the cavity inside), we collect data for both \textbf{Free Flight} and \textbf{Grasping} over a standardized 10-second interaction period for consistent benchmarking. For the \textbf{Free Flight} tests, this period begins after the UAV has had sufficient time to settle in front of the target. We define this as at least 5 seconds at the final setpoint, supplemented by a pilot's visual inspection. For the \textbf{Grasping} tests, the interaction begins at $t_g + 1$s.

\begin{figure}
  \includegraphics[width=0.95\columnwidth]{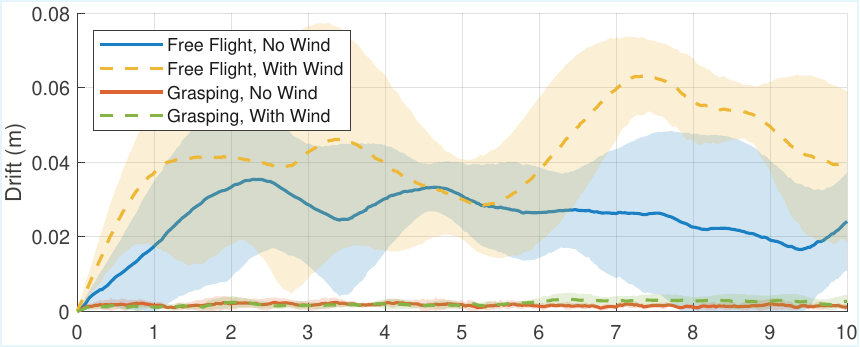}
  \centering
  \vspace{-0.2cm}
  \caption{Transient stabilization performance for the 60\textdegree-pitched tests through quantifying positional drift over time. Each data point is the average of the five trials for that test case, and the shaded region represents 1 SD.}

  \vspace{-0.5cm}
  \label{fig_stability}
\end{figure}

Each test was performed five times. For every grasping trial, the UAV achieved a stable anchor on the first attempt. The results are summarized in Table~\ref{tab:comprehensive_metrics}. Figure~\ref{fig_stability} depicts positional stability across test cases for the more extreme pitched flights, showing mean positional drift over time, with the shaded region representing one standard deviation (SD). 

In the \textbf{Level} tests, the benefit of physical anchoring is immediately apparent. We see an 87\% reduction in the positional drift RMSE in the \textbf{Grasping} configuration compared to the \textbf{Free Flight} baseline. This performance gap widens in \textbf{Pitched} tests because the \textbf{Grasping} performance is unaffected by the attitude change, while the \textbf{Free Flight} positional drift RMSE nearly doubles from 0.015\,m to 0.028\,m. This is clearly visible in Fig.~\ref{fig_stability}, where the \textbf{Free Flight} configuration exhibits noticeable low-frequency oscillations while the grasped UAV maintains a near-zero positional drift.

The benefits of grasping become more critical under wind disturbances. While the \textbf{Free Flight} positional RMSE more than doubles in the \textbf{Level + Wind} scenario, the grasped UAV's performance is largely unaffected, resulting in a 95\% reduction in error compared to the equivalent free-flight performance. This robustness extends to the \textbf{Pitched + Wind} condition, as depicted in Fig.~\ref{fig_stability}.

Overall, physical anchoring ensures consistent, precise performance, with a positional drift RMSE never exceeding 3\,mm across all scenarios. This stability is competitive with similar works in the literature, such as the 4\,mm mean position deviation in~\cite{taylor2024reconfigurable} and the 5\,mm position oscillations in~\cite{hang2019perching}. Crucially, our system maintains this stability across four primary flight conditions, whereas \cite{hang2019perching} and \cite{taylor2024reconfigurable} were only established during level flight.

Despite the stability gains of our anchored system, this comes at the cost of an approximately 35\% increase in energy consumed during the interaction period when compared to \textbf{Free Flight} tests; a trade-off that remains consistent across all \textbf{Grasping} tests. This is primarily due to carrying the additional and offset weight of the gripper. We plan to alleviate this by a lighter gripper design in our future work.

While we omit a full study on pitching downward for brevity, we performed a single grasping test while hovering at a maximum downward pitch angle of 55\textdegree. This yielded similar results to the equivalent maximum upward pitch test, with a positional drift RMSE of 0.002\,m and an angular drift RMSE of 0.31\textdegree, indicating consistent performance across the UAV's range of operation.

\subsection{Gripper Quantification}
\label{sec:grip_quant}

While our experiments thus far demonstrate high stability when anchoring, these benefits are dependent on the grasp quality. Therefore, in this section, we conduct experiments to establish our gripper's operational limits under high interaction forces and ability to adapt to different targets.

\subsubsection{High-Force Tests}

\begin{figure}[t!]
    \centering
    \includegraphics[width=0.85\columnwidth]{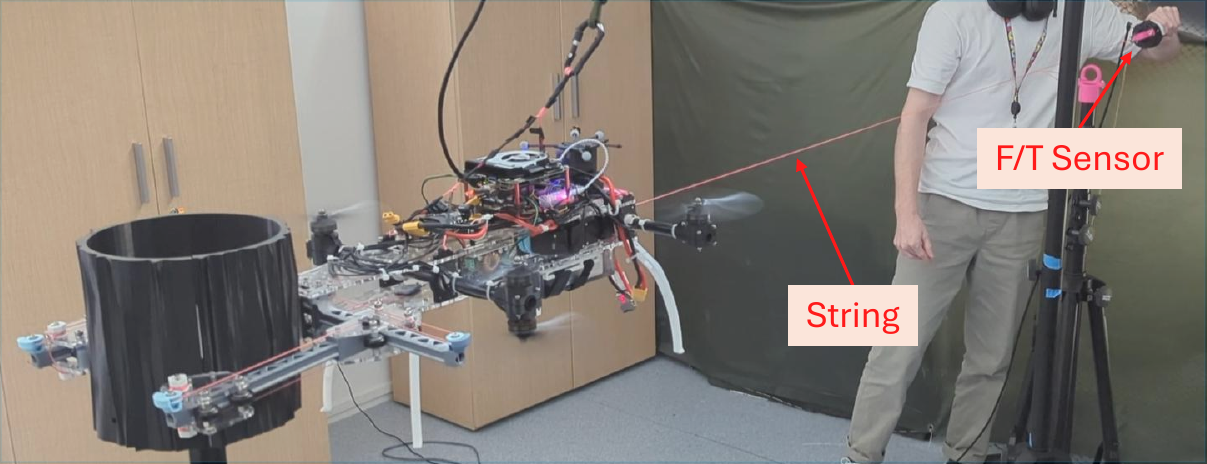}
    \vspace{-0.2cm}
    \caption{Setup for the high-force pulling tests, in which a string is attached to the UAV, aligned with its longitudinal axis. Force is manually applied via the string, and the subsequent tension recorded using a F/T sensor.}
    \vspace{-0.1cm}
    \label{fig_pull_tests}
\end{figure}

\begin{figure}[t!]
    \centering
    \includegraphics[width=0.9\columnwidth]{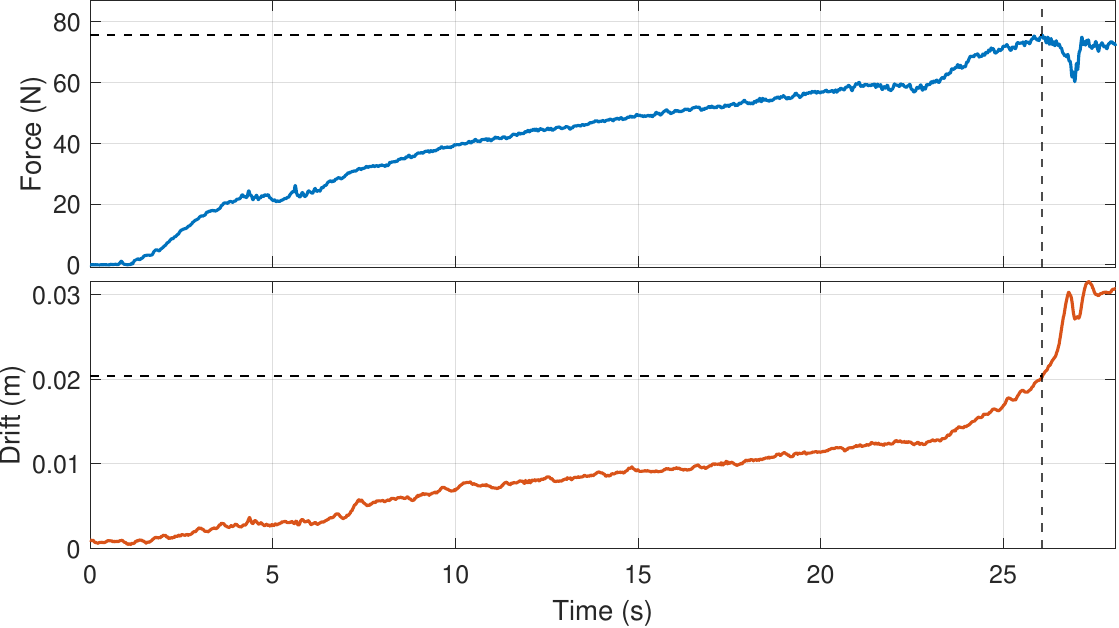}
    \vspace{-0.2cm}
    \caption{\textbf{Maximum Force} test results showing the synchronized force (top) and drift (bottom) relative to the interaction start. Dashed lines denote the failure point where the anchor supports a maximum load of $\approx75\,\text{N}$.}
    \vspace{-0.5cm}
    \label{fig_max_pull_plot}
\end{figure}

With the UAV anchored to the target, we attached a string to the rear of the airframe, aligned approximately with its longitudinal ($X^B$) axis, as shown in Fig.~\ref{fig_pull_tests}. While not fully capturing the forces and torques of tasks like drilling, pulling this string preliminarily tests the UAV's resistance to longitudinal forces common to contact-based tasks. A Bota Systems SensOne force torque (F/T) sensor measured this force during two loading scenarios:
\begin{enumerate}
    \item \textbf{Sustained Loading:} The string is pulled until the UAV reaches a 3D positional drift of $\approx10$\,mm. This force is then maintained for $10$ seconds to evaluate the system's behavior under sustained load. We characterize the performance by residual positional and angular RMSE (calculated relative to the mean of the hold period).
    \item \textbf{Maximum Force:} We apply a ramp force until the anchor fails beyond flex of the structure to determine the maximum force capability of the system.
\end{enumerate}
We ran five \textbf{Sustained Loading} tests, and one \textbf{Maximum Force} test. Table~\ref{tab:pull_tests} summarizes these results, and Fig.~\ref{fig_max_pull_plot} shows the force and drift profiles in \textbf{Maximum Force} test.

Across the five \textbf{Sustained Loading} tests, the targeted 10\,mm positional drift corresponded to an applied force of approximately 41\,N. Throughout the holding period, this force was consistent with a mean RMSE force drift across the five tests of about 1.3\,N relative to the test's mean force. The pose of the UAV was very stable during the holding period with a mean residual positional drift RMSE of 1\,mm.

For the \textbf{Maximum Force} test, the system failed due to sudden reconfiguration of the gripper joints caused by finger pad slippage at $\approx75$\,N, at which point the UAV reached a positional drift of 20\,mm. These results show our platform's ability to remain stable even when exposed to high interaction forces.

\begin{table}[t]
\centering
\caption{Summary of high-force test results. Sustained loading results are reported as the mean $\pm$ 1 sd over five trials.}
\vspace{-0.2cm}
\label{tab:pull_tests}
\begin{tabular}{@{}lc@{}}
\hline
\textbf{Metric} & \textbf{Value} \\
\hline
\multicolumn{2}{@{}l}{\textit{\textbf{Sustained Loading}}} \\
\quad Residual 3D Positional Drift RMSE (m) & 0.001 $\pm$ 0.000 \\
\quad Max Residual Positional Drift (m) & 0.001 $\pm$ 0.000 \\
\quad Residual 3D Angular RMSE ($^{\circ}$) & 0.07 $\pm$ 0.02 \\
\quad Max Residual 3D Angular Drift ($^{\circ}$) & 0.19 $\pm$ 0.07 \\

\multicolumn{2}{@{}l}{\textit{\textbf{Maximum Force}}} \\
\quad Anchor Break Threshold ($N$) & ~75 \\
\hline
\vspace{-0.7cm}
\end{tabular}
\end{table}

\subsubsection{Grasping Versatility}

\begin{figure*}
    \centering
  \includegraphics[width=0.9\textwidth]{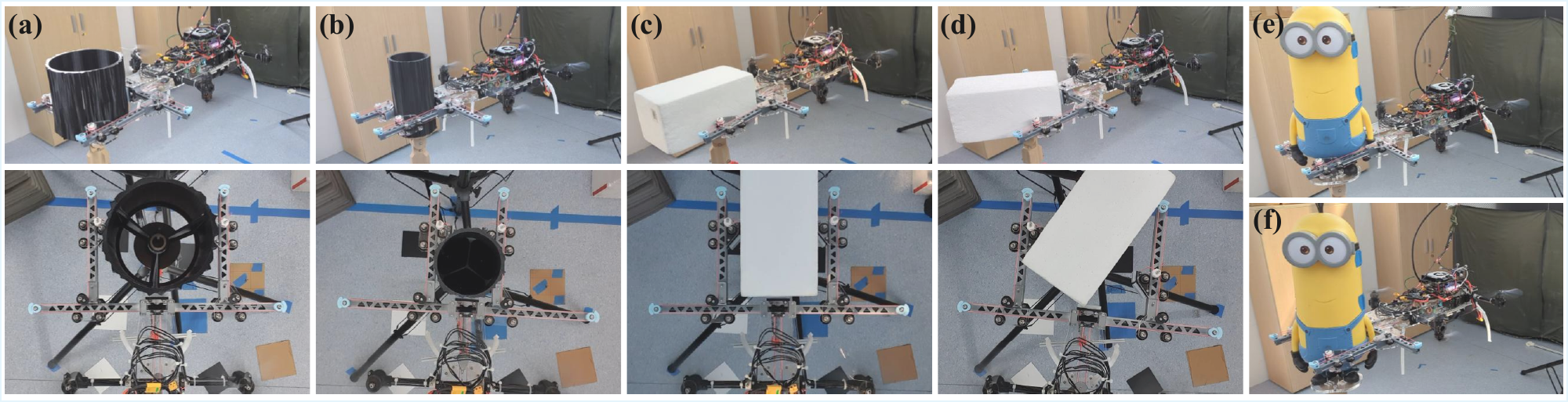}
  \vspace{-0.3cm}
  \caption{Testing our gripper's performance on a range of target objects, including \textbf{(a)} the large branch ($d\approx200$\,mm), \textbf{(b)} a smaller 3D-printed tree branch ($d\approx120$\,mm), \textbf{(c)} a column (front on), \textbf{(d)} a column (angled), \textbf{(e)} a toy figure (around the legs), and \textbf{(f)} a toy figure (around the torso).}
  \vspace{-0.3cm}
  \label{fig_grasp_versatility}
\end{figure*}

We assess the gripper's versatility using \textbf{Level} tests on various targets. These include \textit{tree branches} of various diameters, simple geometries, and complex, asymmetric forms, as shown in Fig.~\ref{fig_grasp_versatility}. For all targets, a secure anchor without visible slippage at $t_g$ was achieved on the first attempt. The results show consistent anchoring across all objects with positional drift RMSE remaining at $\leq$\,3\,mm for every target, and angular drift RMSE remaining at $<$\,0.4\textdegree, comparable to our multi-run baseline established on the \textit{large branch}. This stability held even for the \textit{angled column}, where one finger could not wrap around the target. Here, the intermediate link provided a third contact point to secure a grasp. Overall, our gripper demonstrated versatility across varied geometries and orientations.

\vspace{-0.05cm}

\subsection{Precision Applications}
\label{sec:prec_app}

To qualitatively evaluate the system's ability to maintain a contact point, as required for tasks like drilling, we mounted a felt-tip pen as a dummy end-effector to the front of the UAV to make contact with the target surface. We performed an interaction test using this setup for both the \textbf{Free Flight} and \textbf{Grasping} configurations, under the \textbf{Level + Wind} scenario.

\begin{figure}[t!]
    \centering
    \includegraphics[width=\columnwidth]{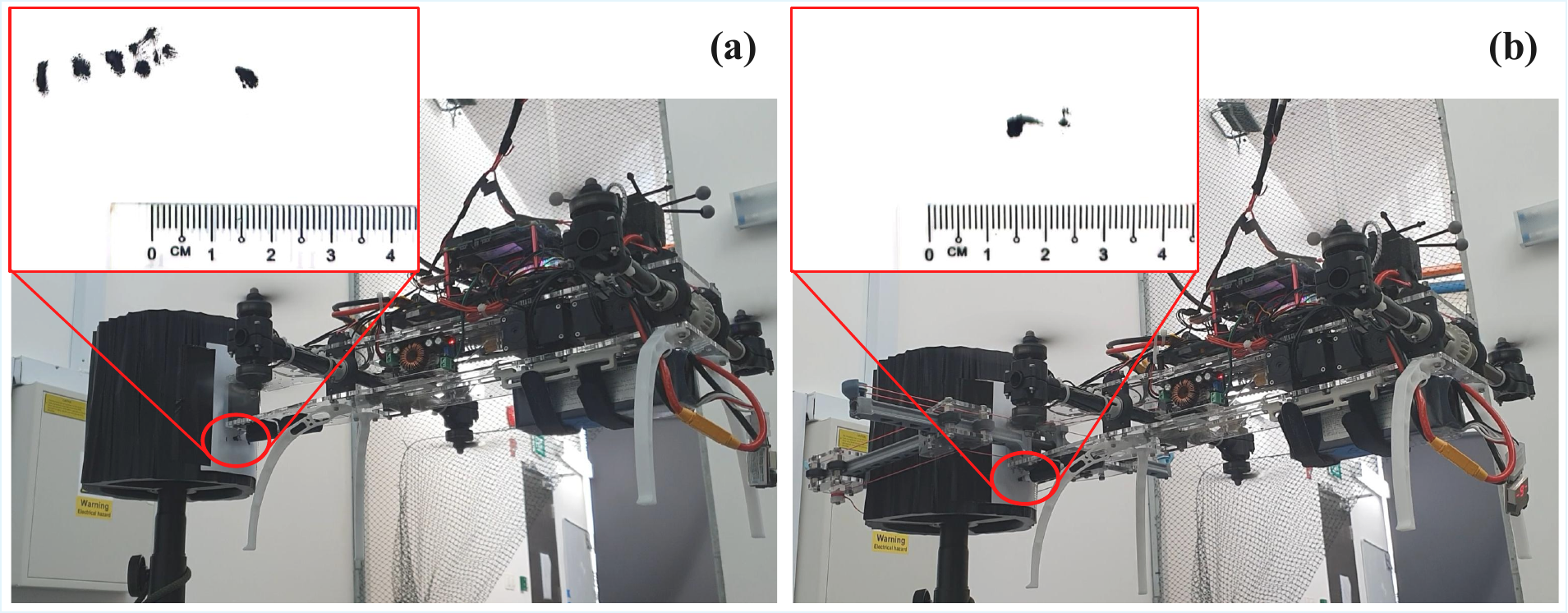}
    \vspace{-0.5cm}
    \caption{End-effector precision under wind disturbances: \textbf{(a)} \textbf{Free Flight} trial results in dispersed pen markings on the target surface; \textbf{(b)} \textbf{Grasping} configuration maintains a concentrated contact point.}
    \vspace{-0.5cm}
    \label{fig_end_effector_test}
\end{figure}

As shown in Fig.~\ref{fig_end_effector_test}, the \textbf{Free Flight} trial resulted in dispersed pen marks separated by up to $\approx35$\,mm across the target due to low-frequency pose oscillations and wind-induced drift. In contrast, the \textbf{Grasping} configuration maintained one contact point throughout the interaction (the second, smaller pen mark was made during initial grasping). This result validates the system's ability to maintain precise contact in environments where free-flight interaction is unfeasible.

\vspace{-0.05cm}
\section{Conclusion}
\label{sec_conclusion}

We introduced a platform that integrates a novel compliant, cable-driven prismatic gripper into a pitch-decoupled tilt-rotor UAV to achieve enhanced stability during physical interaction. Through comprehensive real-robot experiments, we demonstrated that physically anchoring the UAV to a target establishes a significantly more stable work platform compared to the free-flight baseline. This stability holds across ideal, disturbed (windy), and pitched flight conditions. Furthermore, the anchored interface successfully resisted longitudinal reaction forces up to 75\,N and demonstrated versatility across diverse target geometries, all while maintaining the high level of precision required for tasks such as drilling to sample tree health. These results validate our approach of using physical anchoring as a primary means of stabilization, enabling complex, contact-based aerial tasks that would be intractable for conventional multirotors.

Future work includes improving power efficiency via weight reduction, developing \textit{leaning} strategies to offload rotor power, and designing deployable gripper architectures to stow the mass closer to the UAV's center of mass when the gripper is not in use. We will also integrate task-specific end-effectors, like a drill, to evaluate the system's resilience to task-generated forces and validate its real-world viability.

\vspace{-0.02cm}

\section*{ACKNOWLEDGMENT}

GitHub Copilot, powered by Gemini (Google), was used for code auto-completion and troubleshooting within the control software (Sec.~\ref{sec_control}) and data analysis scripts (Sec.~\ref{sec_testing}). Gemini was used to proofread the paper.

Joshua Taylor is an A*STAR scholar under the Singapore International Graduate Award (SINGA) framework.

\bibliographystyle{IEEEtran}
\bibliography{refs.bib}

\end{document}